# Integrative Semantic Dependency Parsing via Efficient Large-scale Feature Selection

**Hai Zhao**                                                     ZHAOHAI@CS.SJTU.EDU.CN
**Xiaotian Zhang**                                               XTIAN.ZH@GMAIL.COM
*Shanghai Jiao Tong University,*
*800 Dongchuan Road, Shanghai,China*

**Chunyu Kit**                                                   CTCKIT@CITYU.EDU.HK
*City University of Hong Kong,*
*Tat Chee Avenue, Kowloon, Hong Kong SAR, China*

## Abstract

Semantic parsing, i.e., the automatic derivation of meaning representation such as an instantiated predicate-argument structure for a sentence, plays a critical role in deep processing of natural language. Unlike all other top systems of semantic dependency parsing that have to rely on a pipeline framework to chain up a series of submodels each specialized for a specific subtask, the one presented in this article integrates everything into one model, in hopes of achieving desirable integrity and practicality for real applications while maintaining a competitive performance. This integrative approach tackles semantic parsing as a word pair classification problem using a maximum entropy classifier. We leverage adaptive pruning of argument candidates and large-scale feature selection engineering to allow the largest feature space ever in use so far in this field, it achieves a state-of-the-art performance on the evaluation data set for CoNLL-2008 shared task, on top of all but one top pipeline system, confirming its feasibility and effectiveness.

## 1. Introduction

The purpose of semantic parsing is to derive the meaning representation for a sentence, usually taking a syntactic parse as input. A popular formalism to represent this kind of meaning is predicate-argument structure and, accordingly, the parsing is to instantiate the predicate and argument(s) in such a structure properly with actual words or phrases from a given sentence. In the context of dependency parsing, it becomes semantic dependency parsing, which takes a syntactic dependency tree as input and outputs a filled predicate-argument structure for a predicate, with each argument word properly labeled with its semantic role in relation to the predicate.

Semantic role labeling (SRL) is one of the core tasks in semantic dependency parsing, be it dependency or constituent based. Conventionally, it is tackled mainly through two subtasks, namely, argument identification and classification. Conceptually, the former determines whether a word is a true argument of a predicate, and the latter what semantic role it plays in relation to the predicate (or which argument it instantiates in a predicate-argument structure). When no predicate is given, two other indispensable subtasks are predicate identification and disambiguation, one to identify which word is a predicate in a sentence and the other to determine the predicate-argument structure for an identified predicate in a particular context.

A pipeline framework was adopted in almost all previous researches to handle these subtasks one after another. The main reason for dividing the whole task of semantic dependency parsing into multiple stages in this way is twofold: maintaining computational efficiency and adopting different favorable features for each subtask. In general, a joint learning system of multiple components is slower than a pipeline system, especially in training. It is also reported by Xue and Palmer (2004) that different features do favor different subtasks of SRL, especially argument identification and classification. The results from the CoNLL shared tasks in 2005 and 2008 (Carreras & Màrquez, 2005; Koomen, Punyakanok, Roth, & Yih, 2005; Surdeanu, Johansson, Meyers, Màrquez, & Nivre, 2008; Johansson & Nugues, 2008) seem to suggest that the pipeline strategy has been the benchmark of technology for the state-of-the-art performance on this specific NLP task.

When most SRL systems are pipeline, an integrated SRL system holds its unique merits, e.g., integrity of implementation, practicality for real applications, a single-stage feature selection benefiting the whole system, an all-in-one model outputting all expected semantic role information, and so on. In particular, it takes into account the interactive effect of features favoring different subtasks and hence holds a more comprehensive view of all features working together as a whole. This article is intended to present our recent research to explore the feasibility of constructing an effective integrated system for semantic dependency parsing that melds all subtasks together into one, including predicate identification/disambiguation and argument identification/classification, for both verbal and nominal predicates, and uses the same feature set for all these subtasks. The core of our research is to verify, through practical implementation and then empirical evaluation, the methodological soundness and effectiveness of this approach. Its success, however, has to be rooted in a solid technical foundation, i.e., a large-scale engineering procedure for efficient mining of effective feature templates from a huge set of feature candidates, a feature space far richer than others ever used before. It is this piece of engineering that brings the potentials of this integrative approach into full play. Another focus of this article is hence to illustrate its technical essentials.

Nevertheless, it is worth pointing out that the term *integrative*, when used in opposite to *pipeline*, can be misleading to mean that all subtasks are carried out jointly in a single run. Instead, it is used to highlight the integrity of our model and its implementation that uses a single representation and feature set to accommodate all these subtasks. Although this approach has its unique advantages in simplifying system engineering and feature selection, the model we have implemented and will present below is not a joint one to accomplish the whole semantic parsing through synchronous determination of both predicates and arguments. These two types of indispensable objects in a semantic parse tree are recognized in succession through decoding using the same trained model.

The rest of the article is organized as follows. Section 2 gives a brief overview of related work, providing the background of our research. Section 4 presents our approach of adaptive pruning of argument candidates to generate head-dependent word pairs for both training and decoding, which underlies the whole process of semantic parsing. The other two key procedures to optimize the parsing, namely, feature selection and decoding, are presented in Section 5 and 6, respectively. The details of evaluation, including evaluation data, experimental results and a comprehensive comparative analysis of the results, are presented in Section 7. Finally, Section 8 concludes our research, highlighting its contributions and the practicality and competitiveness of this approach.

## 2. Related Work

Note that SRL has almost become a surrogate for semantic dependency parsing in the literature of recent years. Most recent research efforts in this field, including the CoNLL shared tasks in 2004 and 2005, have been focused on verbal predicates, thanks to the availability of PropBank (Palmer, Gildea, & Kingsbury, 2005). As a complement to PropBank, NomBank (Meyers, Reeves, Macleod, Szekely, Zielinska, Young, & Grishman, 2004) annotates nominal predicates and their correspondent semantic roles using a similar semantic framework. Although offering more challenges, SRL for nominal predicates has drawn relatively little attention (Jiang & Ng, 2006). The issue of merging various treebanks, including PropBank, NomBank and others, was once discussed in the work of Pustejovsky, Meyers, Palmer, and Poesio (2005). The idea of merging these two treebanks was put into practice for the CoNLL-2008 shared task (Surdeanu et al., 2008). The best system in CoNLL-2008 used two different subsystems to cope with verbal and nominal predicates, respectively (Johansson & Nugues, 2008). Unfortunately, however, there has been no other integrative approach than ours to illustrate a performance so close to that of this system.

In fact, there have been few research efforts in this direction, except a recent one on joint identification of predicates, arguments and senses by Meza-Ruiz and Riedel (2009). They formulate the problem into a Markov Logic Network, with weights learnt via 1-best MIRA (Crammer & Singer, 2003) Online Learning method, and use Cutting Plane Inference (Riedel, 2008) with Integer Linear Programming (ILP) as the base solver for efficient joint inference of the best choice of predicates, frame types, arguments and role labels with maximal *a posteriori* probability. Using CoNLL-2008 data, their system achieves its best semantic $F_1$ 80.16% on the WSJ test set. This is 0.75 percentage point lower than ours, to be reported below, on the whole WSJ+Brown test set. Note that when trained on CoNLL-2008 training corpus, a subset of WSJ corpus, an SRL system has a performance at least 10 percentage points higher on the WSJ than on the Brown test set (Surdeanu et al., 2008).

Both CoNLL-2008 and -2009 shared tasks[1] are devoted to the joint learning of syntactic and semantic dependencies, aimed at testing whether SRL can be well performed using only dependency syntax input. The research reported in this article focuses on semantic dependency parsing. To conduct a valid and reliable evaluation, we will use the data set and evaluation settings of CoNLL-2008 and compare our integrated system, which is the best SRL system in CoNLL-2009 (Zhao, Chen, Kit, & Zhou, 2009), against the top systems in CoNLL shared tasks (Surdeanu et al., 2008; Hajič, Ciaramita, Johansson, Kawahara, Martí, Màrquez, Meyers, Nivre, Padó, Štěpánek, Straňák, Surdeanu, Xue, & Zhang, 2009).[2] Note that these systems achieved higher performance scores in CoNLL-2008 than in CoNLL-2009.

An integrative approach to dependency semantic parsing has its own pros and cons. To deal with its main drawbacks, two key techniques need to be applied for the purpose of efficiency enhancement. One is to bring in auxiliary argument labels that enable further improvement of argument candidate pruning. This significantly facilitates the development of a fast and lightweight SRL system. The other is to apply a greedy feature selection algorithm to perform the task of feature selection from a given set of feature templates. This helps find as many features as possible that are of benefit to the overall process of the parsing. Many individual optimal feature template sets are reported in the literature to have achieved an excellent performance on specific subtasks of SRL. This is the first time that an integrated SRL system is reported to produce a result so close to the state of the art of SRL achieved by those pipelines with individual sub-systems each highly specialized for a specific subtask or a specific type of predicate.

---

1. Henceforth referred to as CoNLL-2008 and -2009, respectively.
2. CoNLL-2008 is an English-only task, while CoNLL-2009 is a multilingual one. Although both use the same English corpus, except some more-sophisticated structures for the former (Surdeanu et al., 2008), their main difference is that semantic predicate identification is not required for the latter.

## 3. System Architecture

Dependencies between words in a sentence, be they syntactic or semantic, can be formulated as individual edges in an abstract graph structure. In practice, a dependency edge has to be built, and its type (usually referred to as its label) to be identified, through proper learning and then decoding. Most conventional syntactic parsing makes use of a property of projectiveness stipulated by the well-formedness of a syntactic tree. In contrast, in dependency parsing, new dependencies have to be built with regard to existing ones. However, this is not the case for semantic parsing, for most semantic parsing results are not projective trees. Instead, they are actually directed acyclic graphs, because the same word can serve as an argument for multiple predicates. Inevitably, a learning model for semantic parsing has to take all word pairs into account when exploring possible dependent relationships.

SRL as a specific task of semantic dependency parsing can be formulated as a word pair classification problem and tackled with various machine learning models, e.g., the Maximum Entropy (ME) model as used by Zhao and Kit (2008). The ME model is also used in this work but only for probability estimation to support the global decoding given below in Section 6, which extends our model beyond a sequential model. Without any constraint, a classifier for this task has to deal with all word pairs in an input sequence and is thus inevitably prone to poor computational efficiency and also unsatisfactory performance. A straightforward strategy to alleviate these problems is to perform proper pruning on both the training sample and test data.

A word pair consists of a word as semantic head and another as semantic dependent, which are conventionally denoted as $p$ (for *predicate*) and $a$ (for *argument*), respectively. We will follow this convention in the feature representation below. Since our approach unifies the two tasks of SRL, namely, predicate identification/disambiguation and argument identification/classification, into one classification framework, there is no need to differentiate between verbal and non-verbal heads, because they are all handled in the same way. This is one of the unique characteristics of our integrated system.

The overall architecture of our system is depicted in Figure 1. An input sentence from a data set in use, be it a training, a development or a test set, is parsed into a word pair sequence by a word pair generator using a pruning algorithm, e.g., the adaptive pruning described below, to eliminate useless pairs. Word pairs so generated from each sentence of the training set are used to train a word pair classifier, which then supports the decoding formulated in Section 6 to search for an optimal set of word pairs from a test sentence to form a semantic parse tree. The decoding first recognizes all predicates in a sentence and then determines the arguments for each predicate by a beam search for their argument role labels. The features used in the classifier are selected from a predefined feature space by a greedy selection procedure using the training and the development set for repeated training and testing to refine a candidate feature set until no more performance gain is achievable (see Section 5). Then the classifier obtained this way with the selected features is tested on the test set.

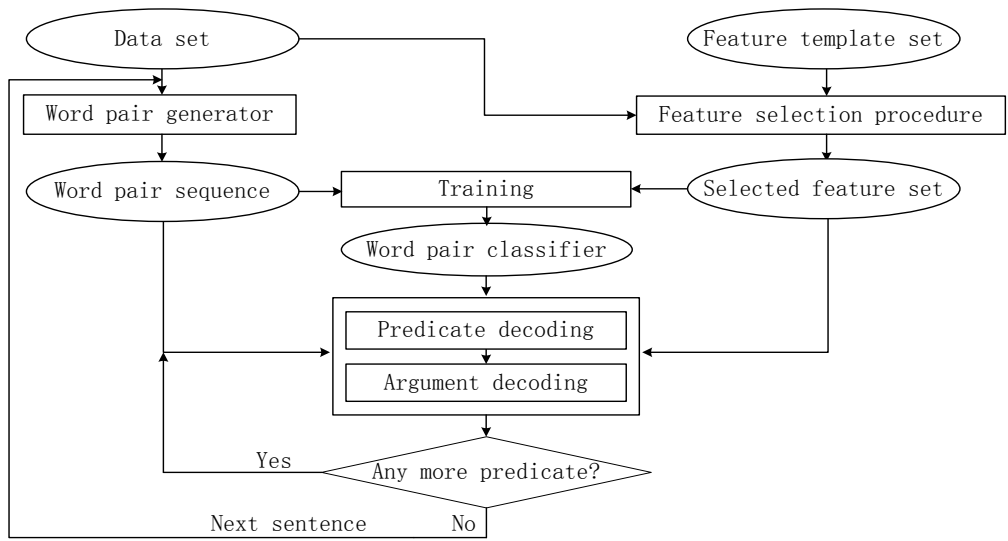

Figure 1: Illustration of system architecture and work flow of training and testing

## 4. Adaptive Argument Pruning

Word pairs are derived from a sentence for the classifier in the following ways. (1) For predicate identification/disambiguation, each word pair consists of the virtual root (VR) of a semantic parse tree under construction (whose root is virtually preset), as head, and a predicate candidate as its dependent. Theoretically, all words in the sentence in question can be a predicate candidate. To reduce their number, we opt for a simple POS tag pruning strategy that only verbs and nouns are allowed as predicate candidates. (2) For argument identification/classification, each word pair consists of an identified predicate, as head, and another word as its dependent (or its argument, in conventional term). Potentially, any other word in the same sentence can be its argument candidate. Pruning off as many argument candidates as possible is thus particularly significant in improving the efficiency and performance of the classifier.

There are two ways to collect argument candidates for a given predicate, one from the syntactic dependency tree and the other from the linear path of an input sentence. For the former (referred to as *synPth* hereafter), we use a dependency version of the pruning algorithm by Xue and Palmer (2004), which is given as follows with a necessary modification to allow a predicate itself also to be included in its own argument candidate list, because a nominal predicate sometimes takes itself as its own argument.

| ID | FORM[a] | LEMMA | POS | HEAD[b] | DEPREL[c] | PRED[d] | | ARG Label[e] | |
|----|---------|-------|-----|---------|-----------|---------|---|-----------|---|
| 1 | Investor | investor | NN | 2 | NMOD | _ | A0 | _ | _ |
| 2 | focus | focus | NN | 3 | SBJ | focus.01 | _ | A1 | _ |
| 3 | shifted | shift | VBD | 7 | OBJ | shift.01 | _ | _ | A1 |
| 4 | quickly | quickly | RB | 3 | MNR | _ | _ | AM-MNR | _ |
| 5 | , | , | , | 7 | P | _ | _ | _ | _ |
| 6 | traders | trader | NNS | 7 | SBJ | _ | _ | _ | A0 |
| 7 | said | say | VBD | 0 | ROOT | say.01 | _ | _ | _ |
| 8 | . | . | . | 7 | P | _ | _ | _ | _ |

a. Word form, or token.
b. Syntactic head of the current token, identified by an ID.
c. Syntactic dependency relation of the current token to its HEAD.
d. Roleset of a semantic predicate.
e. Argument labels for semantic predicates in text order.

Table 1: An example of input sentence from CoNLL-2008 shared task data set

Initialization: Given a predicate as the current node in a syntactic dependency tree.

1. Collect all its syntactic children as argument candidates, by traversing the children from left to right.

2. Reset the current node to its syntactic head and repeat Step 1 till the root of the tree.

3. Collect the root and stop.

This algorithm is effective in collecting both words in the path from a given predicate to the root and their children as argument candidates. However, a more efficient one is still needed to lend stronger support to our SRL system that is designed to tackle argument identification/classification in a single stage. Following the observation that arguments usually tend to surround their predicate in a close distance, the auxiliary label `noMoreArg` is introduced to signify where the pruning stops collecting argument candidates. For training sample generation, this label is assigned to the next word as soon as the arguments of the current predicate have been saturated with previously collected words, in light of the original training data as illustrated in Table 1. Accordingly, the pruning process stops collecting any more candidates. For decoding, it signals the decoder to stop searching, along a similar traverse as the pruning, for any more arguments for an identified predicate. This adaptive technique improves the pruning efficiency significantly, saving about 1/3 training time and memory at the cost of missing very few more true arguments than the pruning without this label, according to our experiments. The training sample generated this way from the sentence in Table 1, by means of both POS pruning and the above pruning algorithm, is illustrated in Table 2, with a few class labels in the third column.

To collect argument candidates along the linear path (referred to as *linPth* hereafter) instead of the syntactic tree of a sentence, the classifier will search through all words around a given predicate. In a way similar to how the pruning along *synPth* is improved, two auxiliary labels, namely, `noMoreLeftArg` and `noMoreRightArg`, are introduced to signify where the adaptive pruning along *linPth* stops, skipping those words too far away from the predicate. Given below is an example to illustrate how these two labels are used, where *e* in the input sequence is a predicate with two arguments, labeled with A0 and A1, respectively. The two labels are assigned to the next two words $c$ and $g$, respectively, indicating no more arguments farther than them from the predicate. Accordingly, the word sequence from $c$ to $g$ are taken as training sample.

| Head-dependent word pair | | Label |
|---|---|---|
| _VR | Investor | `NONE_PRED` |
| _VR | focus | `01` |
| _VR | shifted | `01` |
| _VR | traders | `NONE_PRED` |
| _VR | said | `01` |
| focus | Investor | `A0` |
| focus | focus | `noMoreArg` |
| shifted | focus | `A1` |
| shifted | quickly | `AM-MNR` |
| shifted | said | `noMoreArg` |
| said | shifted | `A1` |
| said | , | `NONE_ARG` |
| said | traders | `A0` |
| said | . | `NONE_ARG` |

Table 2: An example of training sample generated via pruning

```
a   b        c            d   e   f           g            h   .
_   _   noMoreLeftArg  A1   _   A0  noMoreRightArg  _   _
```

The total list of class labels in our model, including those from the CoNLL-2008 data set and a few auxiliary ones newly introduced on purpose, is provided in Table 9 in Appendix A. These labels are in three categories, namely, 22 PropBank sense labels as predicate classes, 54 argument classes, and 2–3 auxiliary labels as extra classes, for a total of 78-79. Pruning along *linPth* needs one more label than that along *synPth*. Note that our work does not assume whether the same sense label in the training and the test set means the same for different words. The tendency of a particular word form to associate with its senses in a statistically significant way throughout the data set allows our classifier to predict sense labels using word form features.

In principle, an auxiliary label is assigned to the last item in the sample that is generated for a predicate via pruning along a traversal order, be it syntactic or linear. That is, it is assigned to the first item immediately after the last argument of the predicate has been seen during the pruning. An auxiliary label is treated in exactly the same way as all other argument labels during training and decoding, except its extra utility to signal where to stop a search.

## 5. Feature Generation and Selection

Following many previous works (Gildea & Jurafsky, 2002; Carreras & Màrquez, 2005; Koomen et al., 2005; Màrquez, Surdeanu, Comas, & Turmo, 2005; Dang & Palmer, 2005; Pradhan, Ward, Hacioglu, Martin, & Jurafsky, 2005; Toutanova, Haghighi, & Manning, 2005; Jiang & Ng, 2006; Liu & Ng, 2007; Surdeanu, Marquez, Carreras, & Comas, 2007; Johansson & Nugues, 2008; Che, Li, Hu, Li, Qin, Liu, & Li, 2008), we carefully examine the factors involved in a wide range of features that have been or can be used to facilitate the undertaking of the two SRL subtasks, for both verbal and nominal predicates. Our endeavor is to further decompose these factors into some more fundamental elements, so that the largest possible space of feature templates can be explored for more effective and novel combinations of them into features.

## 5.1 Feature Element

All features adopted for this work are intended to make full use of these elements, which are mainly drawn from the *word property* and *syntactic connection* of a node in the syntactic parse tree of an input sentence. The sequences or sets of tree nodes, whose basic elements are drawn to form features via feature generation by means of many predefined feature templates, are identified through the *path* and *family* relations as stipulated below.

**Word Property**  This type of elements include word form (denoted as `form` and its split form as `spForm`),[3] lemma (as `lemma` and `spLemma`), part-of-speech tag (as `pos` and `spPos`), and syntactic and semantic dependency labels (as `dprel` and `semdprel`).[4]

**Syntactic Connection**  This includes syntactic head (as `h`), left/right farthest/nearest child (as `slm`, `ln`, `rm` and `rn`), and high/low support verb or noun. Note that along the path from a given word to the root of a syntactic tree, the first/last verb is called its low/high support verb, respectively. This notion is widely adopted in the field (Toutanova et al., 2005; Xue, 2006; Jiang & Ng, 2006).[5]  In this work, we extend it to both nouns and prepositions. Besides, we also introduce another syntactic head feature `pphead` for a given word in question, to retain its left most sibling if headed by a preposition, or its original head otherwise, aimed at drawing utility from the fact that a preposition usually carries little semantic information. The positive effect of this new feature is confirmed by our experiments.

**Path**  There are two basic types of path from an argument candidate $a$ to a given predicate $p$, namely, the linear path `linePath` as the sequence of input words between them (inclusive) and the other path `dpPath` between them (inclusive) as in their syntactic dependency tree. Given the two paths from them to the root $r$ of the tree that meet at a node $r'$, we have their common part `dpPathShare` from $r'$ to $r$, their different parts `dpPathArgu` and `dpPathPred` from $a$ and $p$ to $r'$, respectively, and the path `dpPath` between $a$ and $p$. Similarly, we have a `dpPath` between any two nodes in a syntactic tree.

**Family**  Two child sets are differentiated for a given predicate or argument candidate, one (as `children`) including all syntactic children and the other (as `noFarChildren`) excluding only the leftmost and the rightmost one. The latter is introduced as a feature to differentiate the modifiers (i.e., children) close to the head from those far away.

---

3. Note that in CoNLL-2008, many treebank tokens are split at the position of a hyphen (-) or a forward slash (/), resulting in two types of form for each, namely, non-split and split.

4. The `lemma` and `pos`, for both training and test, are directly from the pre-analyzed columns of an input file, automatically generated by the organizer of CoNLL shared tasks.

5. Note that the notion of the term *support verb* is slightly different in these works. It is used here to refer to a verb that introduces a long-distance argument to a nominal predicate from outside of the noun phrase headed by the nominal predicate.

**Others** There are also a number of other elements, besides those in the above categories, that play a significant role in feature generation. Many of them are derived from inter-word relationships. Listed below are a number of representative ones.

**dpTreeRelation** It returns the relationship of $a$ and $p$ in an input syntactic tree. The possible values for this feature include **parent**, **sibling**, etc.

**isCurPred** It checks whether a word in question is the current predicate, and returns the predicate itself if yes, or a default value otherwise.

**existCross** It checks if a potential dependency relation between a given pair of words may cross any existing relation in the semantic tree under construction.

**distance** It returns the distance between two words along a given path, be it **dpPath** or **linePath**, in number of words.

**existSemdprel** It checks whether a given argument label under a predicate has been assigned to any other word.

**voice** It returns either **Active** or **Passive** for a verb and a default value for a noun.

**baseline** A small set of simple rules[6] are used to generate SRL output as the baseline for CoNLL evaluation (Carreras & Màrquez, 2005). This baseline output can be selectively used as features, in two categories: **baseline_A**$x$ tags the head of the first NP before and after a predicate as A0 and A1, respectively, and **baseline_Mod** tags the modal verb dependent of a predicate as AM-MOD.

A number of features such as **existCross** and **existSemdprel** have to depend on the semantic dependencies or dependency labels in the existing part of a semantic parse tree under (re)construction for a sentence, be it for training or decoding. Note that both training and decoding first take the candidate word pairs from a given sentence as input, as illustrated in Table 2, and then undergo a process of selecting a subset of the candidates to (re)construct a semantic parse tree, which consists of a root, some predicate(s) as its child(ren), and the argument(s) of the predicate(s) as its grandchild(ren). The decoding infers an optimal semantic tree for a sentence with the aid of a trained ME model (see Section 6). The training reconstructs the gold standard semantic tree of an input sentence when scanning through its word pairs in sequence and differentiating the true ones in the tree from the others. The true ones rebuild the tree part by part. All features (including **existCross** and **existSemdprel**) extracted from both the true ones, as in the partially (re)built parts of the tree, and the others in the current context are fed to the ME model for training. In other words, the feature generation is based on gold standard argument labels during training and on predicted ones during decoding.

## 5.2 Feature Generation

Sequences of syntactic tree nodes are first collected by means of the paths and/or the family relations defined above. Three strategies are then applied to combine elements of the same type (e.g., **form**, **spPos**) from these nodes into a feature via string concatenation. The three strategies of concatenation are: (1) sequencing (as **seq**), which concatenates given element strings in their original order in the path, (2) unduplicating (as **noDup**), which further frees

---

`seq` from adjacent duplicates, and (3) bagging (as `bag`), which concatenates unique element strings in alphabetical order.

Given below are a number of typical feature templates to illustrate how individual features are derived in the ways as described above, with the aid of the following operators: `x+y` (the concatenation of `x` and `y`), `x.y` (the attribute `y` of `x`), `x:y` (the path from `x` to `y`), and `x:y|z` (the collection of all instances of attribute `z` along the path from `x` to `y`).

`a.lm.lemma`  The lemma of the leftmost child of the argument candidate `a`.

`p.h.dprel`  The dependency label of the syntactic head of predicate candidate `p`.

`p₋₁.pos + p.pos`  The concatenation of the POS tags of two consecutive predicates.

`a:p|dpPath.lemma.bag`  The `bag` of all lemmas along the `dpPath` from `a` to `p`.

`a:p.highSupportNoun|linePath.dprel.seq`  The `seq` of all dependency labels along the `linePath` from `a` to the high support noun of `p`.

In this way, a set of 781 feature templates,[7] henceforth referred to as $FT$, is generated to specify the allowable feature space for feature selection. Many of them are generated by analogy to existing feature templates in the literature. For example, given a feature template like `a.lm.lemma` which has been used in some previous works, its analogous ones such as `a.rm.lemma`, `a.rn.lemma` and `a.ln.lemma` are included in the $FT$.

Predicate sense labels in the data set are also utilized as a type of element in various feature templates in the $FT$. However, it is worth noting that the same sense label associated with different words, e.g., `02` in `take.02` and in `say.02`, is not assumed to have anything in common or anything to do with each other. For predicate disambiguation, however, these features always combine a predicate sense with a word form, and hence naturally differentiate between the same sense label for different words. To predict a predicate sense label is always to predict it in association with a word form. That is, a sense label is never used in separation from a word form. In this way, our model gives a very high precision for sense label prediction according to our empirical results.

## 5.3 Feature Template Selection

It is a complicated and hence computationally expensive task to extract an optimal subset of feature templates from a large feature space. For the sake of efficiency, a greedy procedure for feature selection has to be applied towards this goal, as illustrated in many previous works, e.g., by Jiang and Ng (2006), and Ding and Chang (2008). The algorithm that we implemented for this purpose is presented in Algorithm 1 below, which imposes fewer assumptions than those in previous works, aiming at a higher efficiency. It repeats two main steps until no further performance gain is achievable on the given development set:

1. Include any template from the rest of $FT$ into the current set of candidate templates if its inclusion would lead to a performance gain.

2. Exclude any template from the current set of candidate templates if its exclusion would lead to no deterioration in performance. By repeatedly adding/removing the most/least useful template, the algorithm aims to return a better or smaller candidate set for next round.

---

7. Available at http://bcmi.sjtu.edu.cn/~zhaohai/TSRLENAllT.txt, in a macro language as used in our implementation, far not as readable as the notation of the illustrations given here.

Given $n$ candidate feature templates, the algorithm by Ding and Chang (2008) requires $O(n^2)$ time to execute a training/test routine, whereas the one by Jiang and Ng (2006) requires $O(n)$ time, assuming that the initial set of feature templates is "good" enough and the others can be handled in a strictly incremental way. The time complexity of our algorithm can also be analyzed in terms of the execution time of the training-and-test routine $scr(M(.))$, for all other subroutines such as sorting are negligible while compared against its execution time. In Algorithm 1, RECRUITMORE first calls this routine $|FT - S| \leq n$ times in the *for* loop, and then SHAKEOFF calls it $|S_{\max}| \leq n$ times to prepare for the sorting, followed by at most another $|S_{\max}|$ times in the inner *while* loop. Assuming that the first *while* loop and the outer *while* in SHAKEOFF iterate $k_1$ and $k_2$ times, respectively, the algorithm is of $O(k_1(|FT - S| + k_2(|S_{\max}| + |S_{\max}|))) = O(k_1 k_2 n)$ time.

Empirically, however, we have $k_1, k_2 << n$, in that our experiments seldom show any $k_1 > 5$ or $k_2 > 10$, especially when running with $1/10$ $FT$ randomly chosen as the initial $S$. In particular, the first *while* loop often iterates only 2-3 times, and after its first iteration $k_2$ drops rapidly. The observation that $k_1 k_2$ varies only in a very limited range suggests that we may have $O(k_1 k_2 n) = O(n)$ as an empirical estimation of the efficiency of the algorithm in this particular context. A reasonable account for this is that as the first *while* loop comprises of only two functions, namely, RECRUITMORE to recruit positive feature templates and SHAKEOFF to filter out negative ones, so as to improve the model in either case, it is likely that the positive/negative ones remain positive/negative consistently throughout the looping. As a result, only very few of them remain outside/inside the candidate set for further recruiting/filtering after a couple of iterations of the loop.

This efficiency allows a large-scale engineering of feature selection to be accomplished at a reasonable cost of time. In our experiments with $1/10$ $FT$ randomly selected as the initial $S$, the greedy selection procedure was performed along one of the two argument candidate traverse schemes (i.e., the *synPth* and *linPth*) on NomBank, PropBank or their combination, and output six feature template sets $S_N^s$, $S_P^s$, $S_{N+P}^s$, $S_N^l$, $S_P^l$ and $S_{N+P}^l$, of 186, 87, 246, 120, 80 and 118 selected templates, respectively, for performance evaluation and comparison. About 5500 machine learning routines ran for the *synPth* scheme and nearly 7000 routines for the *linPth*. A contrastive analysis of these template sets, with a focus on the top 100 or so most important templates from each of them, is presented in Appendix A through Tables 9-17, where the rank columns present the rankings of feature templates in terms of their importance in respective feature template sets. The importance of a feature template in a template set is measured in terms of the performance change by adding or removing that template, and the performance of a model using a template set is measured by its labeled $F_1$ score on a given test set, following the conventional practice of SRL evaluation in CoNLL shared tasks.

It is interesting to note that the six template sets have a tiny intersection of only 5 templates, as listed in Table 10, each manifesting a notable variance of importance ranking in different sets. Excluding these five, the rest of the overlap of the top 100 of the *synPth* sets $S_N^s$, $S_P^s$ and $S_{N+P}^s$ is also very small, of only 11 templates, in contrast to that of the *linPth* sets $S_N^l$, $S_P^l$ and $S_{N+P}^l$, which is about 4 times larger, of 46 templates; as listed in

---

**Algorithm 1** Greedy Feature Selection

---

**Input**

A training data set: $T$

A development data set: $D$

The set of all feature templates: $FT$

**Denotation**

$M(S) = M(S, T)$, a model using feature template set $S$, trained on $T$;

$scr(M) = scr(M, D)$, the evaluation score of model $M$ on $D$;

Since $T$ and $D$ are fixed, let $scr(M(S)) = scr(M(S, T), D)$ for brevity.

**Algorithm**

1: $S = \{f_0, f_1, ..., f_k\}$, a random subset of $FT$;          ▷ $FT$: a globally accessible constant
2: **while  do**
3:     $C_r = \text{RECRUITMORE}(S)$;
4:     **if** $C_r == \{\}$ **then return** $S$;
5:     $S' = \text{SHAKEOFF}(S + C_r)$;
6:     **if** $scr(M(S)) \geq scr(M(S'))$ **then return** $S$;
7:     $S = S'$;
8: **end while**

1: **function** RECRUITMORE$(S)$          ▷ *Retrieve more positive templates from $FT - S$*
2:     $C_r = \{\}$, and $p = scr(M(S))$;
3:     **for** each $f \in FT - S$ **do**
4:         **if** $p < scr(M(S + \{f\}))$ **then** $C_r = C_r + \{f\}$;
5:     **end for**
6:     **return** $C_r$;
7: **end function**

1: **function** SHAKEOFF$(S_{\max})$          ▷ *Shake off useless templates from $S_{\max}$*
2:     **while  do**
3:         $S = S_0 = S_{\max}$;
4:         sort $S$ in the descending order[a] of $scr(M(S - \{f\}))$ for each $f \in S$;
5:         **while** $(S = S - \{f_0\}) \neq \{\}$ **do**
6:             $S_{\max} = \text{argmax}_{x \in \{S_{\max}, S\}} scr(M(x))$;          ▷ *Drop $f_0 \in S$ if it is useless*
7:         **end while**
8:         **if** $S_0 == S_{\max}$ **then return** $S_0$;          ▷ *If none dropped*
9:     **end while**
10: **end function**

---

a. Namely in the ascending order of the importance of $f$ in $S$, estimated by $scr(M(S)) - scr(M(S - \{f\}))$.

Tables 11 and 12, respectively. Besides these shared templates, these six sets hold 84, 71, 84, 69, 29 and 67 others in their top 100, as listed in Tables 13-18, respectively, where a negative/positive subscript denotes a preceding/following word. For example, `a.lm_-1.lemma` returns the lemma of the previous word of `a`'s left most child.

The rather small overlap of the six sets suggests that the greedy feature selection algorithm maintains a stable efficiency while working out these template sets of huge divergence, lending evidence to support the empirical estimation above. Despite this divergence, each of these template sets enables our SRL model to achieve a state-of-the-art performance on the CoNLL-2008 data set,[8] indicating the effectiveness of this approach, for which more details of evaluation will be provided in Section 7 below.

## 6. Decoding

Following exactly the same procedure of generating the training sample, our ME classifier, after training, outputs a series of labels for the sequence of word pairs generated from an input sentence, inferring its predicates and their arguments one after another. Different from most existing SRL systems, it instantiates an integrative approach that conducts all predication with the same trained model. However, following the common practice of incorporating task-specific constraints into a global inference (Roth & Yih, 2004; Punyakanok, Roth, Yih, & Zimak, 2004), we opt for further developing a decoding algorithm to infer the optimal argument structure for any predicate that is identified this way by the classifier. The main differences of our work from Punyakanok et al. (2004) are that (1) they use ILP for joint inference, which is exact, and we use beam search, which is greedy and approximate, and (2) the constraints (e.g., no duplicate argument label is allowed) that they impose on arguments through individual linear (in)equalities are realized through our constraint fulfillment features (e.g., `existCross` and `existSemdprel`).

Specifically, the decoding is to identify the arguments among candidate words by inferring the best semantic role label for each candidate (*cf.* the training sample in Table 2 with one label per word). Let $A = \{a_0, a_1, ..., a_{n-1}\}$ be the candidates for a predicate, where each $a_i$ embodies all available properties of a word, including a candidate label, and let $A_i' = a_0\, a_1\, ... \, a_{i-1}$ be a partial argument structure (of our target under search) that has been determined and ready for use as the context for inferring the next argument. Instead of counting on best-first search, which simply keeps picking the next best argument according the conditional probability $p(a_i|A_i')$, we resort to a beam search for a better approximation of the global optimization for the maximal probability in

$$\tilde{A} = \operatorname*{argmax}_{A' \subseteq A} \prod_{i=0}^{n} p(a_i|A_i'),\qquad(1)$$

where $A_i'$ consists of the first $i$ elements of $A'$. Ideally, the beam search returns the most probable subset of $A$ as arguments for the predicate in question. It rests on a conditional maximum entropy sequential model incorporating global features into the decoding to infer the arguments that are not necessarily in a sequential order. As in previous practice, our ME model adopts a tunable Gaussian prior (Chen & Rosenfeld, 1999) to estimate $p(a_i|A_i')$ and applies the L-BFGS algorithm (Nocedal, 1980; Nash & Nocedal, 1991) for parameter optimization.

---

8. Note that an early version of this model also illustrated a top-ranking performance on CoNLL-2009 multilingual data sets (Zhao, Chen, Kit, & Zhou, 2009).

## 7. Evaluation

The evaluation of our SRL approach is conducted with various feature template sets on the official training/development/test corpora of CoNLL-2008 (Surdeanu et al., 2008). This data set is derived by merging a dependency version of the Penn Treebank 3 (Marcus, Santorini, & Marcinkiewicz, 1993) with PropBank and NomBank. Note that CoNLL-2008 is essentially a joint learning task on both syntactic and semantic dependencies. The research presented in this article is focused on semantic dependencies, for which the primary evaluation measure is the semantic labeled $F_1$ score (Sem-$F_1$). Other scores, including the macro labeled $F_1$ score (Macro-$F_1$), which was used to rank the participating systems in CoNLL-2008, and Sem-$F_1$/LAS, the ratio between labeled $F_1$ score for semantic dependencies and the labeled attachment score (LAS) for syntactic dependencies, are also provided for reference.

### 7.1 Syntactic Input

Two types of syntactic input are used to examine the effectiveness of our integrative SRL approach. One is the gold standard syntactic input available from the official data set and the other is the parsing results of the same data set by two state-of-the-art syntactic parsers, namely, the MSTparser[9] (McDonald, Pereira, Ribarov, & Hajič, 2005; McDonald & Pereira, 2006) and the parser of Johansson and Nugues (2008). However, instead of using the original MSTparser, we have it substantially enriched with additional features, following Chen, Kawahara, Uchimoto, Zhang, and Isahara (2008), Koo, Carreras, and Collins (2008), and Nivre and McDonald (2008). The latter one, henceforth referred to as J&N for short, is a second-order graph-based dependency parser that takes advantage of pseudo-projective techniques and resorts to syntactic-semantic reranking for further refining its final outputs. However, only its 1-best outputs before the reranking are used for our evaluation, even thought the reranking can slightly improve its parsing performance. Note that this reward of reranking through joint-learning for syntactic and semantic parsing is gained at a huge computational cost. On the contrary, our approach is intended to show that highly comparable results can be achieved at much lower cost.

### 7.2 Experimental Results

The effectiveness of the proposed adaptive approach to pruning argument candidates is examined with the above three syntactic inputs, and the results are presented in Table 3,[10] where a coverage rate is the proportion of true arguments in pruning output. Note that using auxiliary labels does not affect this rate, which has to be accounted for by the choice of traverse path and the quality of syntactic input, as suggested by its difference in the *synPth* rows. The results show that the pruning reduces more than 50% candidates along

---

9. Available at http://mstparser.sourceforge.net.
10. Decimal figures in all tables herein are percentages unless otherwise specified.

| Syntactic Input (LAS) | Path | Original | Pruning | Reduction | Coverage |
|---|---|---|---|---|---|
| MST (88.39%) | *linPth* | 5.29M | 1.57M | -70.32 | 100.0 |
| | *synPth* | 2.15M | 1.06M | -50.70 | 95.0 |
| J&N (89.28%) | *linPth* | 5.28M | 1.57M | -70.27 | 100.0 |
| | *synPth* | 2.15M | 1.06M | -50.70 | 95.4 |
| Gold (100.0%) | *linPth* | 5.29M | 1.57M | -70.32 | 100.0 |
| | *synPth* | 2.13M | 1.05M | -50.70 | 98.4 |

Table 3: Reduction of argument candidates by the adaptive pruning

| Path $x$ | $S_N^x$ | $S_P^x$ | $S_{N+P}^x$ |
|---|---|---|---|
| *linPth* | 7,103 | 7,214 | 7,146 |
| *synPth* | 5,609 | 5,470 | 5,572 |
| Reduction | -21.03 | -24.18 | -22.03 |

Table 4: Number of executions of the training-and-test routine in greedy feature selection

*synPth*, at the cost of losing 1.6-4.6% true ones, and more than 70% along *linPth* without any loss. Nevertheless, the candidate set so resulted from *synPth* is 1/3 smaller in size than that from *linPth*.

The number of times that the training-and-test routine is executed in the greedy selection of all six feature sets are presented in Table 4, showing that *synPth* saves 21%-24% execution times. Given the estimation of the time complexity of the selection algorithm as $O(k_1 k_2 n)$ for executing the routine, empirically we have $7 < k_1 k_2 < 10$ on a feature space of size $n = 781$ for our experiments, verifying the very high efficiency of the algorithm.

As pointed out by Pradhan, Ward, Hacioglu, Martin, and Jurafsky (2004), argument identification (before classification) is a bottleneck problem in the way of improving SRL performance. Narrowing down the set of predicate candidates as much as possible in a reliable way has been shown to be a feasible means to alleviate this problem. The effectiveness of our adaptive pruning for this purpose can be examined through comparative experiments in terms of time reduction and performance enhancement. The results from a series of such experiments are presented in Table 5, showing that the adaptive pruning saves the training and test time by about 30% and 60%, respectively, while enhancing the performance (in Sem-$F_1$ score) by 23.9%–24.8%, nearly a quarter. These results also confirm a significant improvement upon its non-adaptive origin (Xue & Palmer, 2004) and the twofold benefit of pruning off arguments far away from their predicates, which follows from the assumption that true arguments tend to be close to their predicates. It is straightforward that using the `noMoreArg` label reduces more training samples than not using (see Section 4) and hence leads to a greater reduction of training time. Using this label also decreases the test time remarkably. During decoding, a `noMoreArg` label, once assigned a probability higher than all other possible role labels for the current word pair, hints the decoder to stop working on the next word pair, resulting in a further test time reduction by 18.5–21.0 percentage points upon the non-adaptive pruning. The particularly low performance without pruning also reflects the soundness of the motivation for candidate pruning from both

| Bank | Features | Pruning[a] | Training | Redu. | Test | Redu. | Sem-$F_1$ |
|---|---|---|---|---|---|---|---|
| | | − | 122,469s | | 747s | | 66.85 |
| PropBank | 87 | −Adaptive[b] | 109,094s | -10.9 | 372s | -50.2 | 80.59 |
| | | +Adaptive | 83,208s | -32.1 | 234s | -68.7 | 82.80 |
| NomBank | | − | 432,544s | | 2,795s | | 64.85 |
| + | 246 | −Adaptive | 392,216s | -9.3 | 1,615s | -42.2 | 79.77 |
| PropBank | | +Adaptive | 305,325s | -29.4 | 1,029s | -63.2 | 80.91 |

a. Syntactic input: MST; Traverse scheme: *synPth*; Machine configuration: Four six-core Intel® Xeon® X5690 3.46GHz processors, 48G memory.
b. The original pruning as in Xue and Palmer (2004), not using `noMoreArg`.

Table 5: Time reduction and performance enhancement by the adaptive pruning

| Syn. Input (LAS) | Feature Set | Path $x$ | Nomi- $F_1{}_N^x$ | Verb- $F_1{}_P^x$ | Nomi- $F_1{}_{N+P}^x$ | Verb- $F_1{}_{N+P}^x$ | Sem- $F_1{}_{N+P}^x$ | Sem- $F_1{}_{N+P}^x$/LAS |
|---|---|---|---|---|---|---|---|---|
| MST (88.39%) | Initial | *linPth* | 44.58 | 58.83 | 41.18 | 56.34 | 51.14 | 57.86 |
| | | *synPth* | 44.67 | 63.24 | 42.42 | 61.28 | 54.79 | 61.99 |
| | Selected | *linPth* | 77.93 | 82.72 | 76.75 | 82.30 | 80.05 | 90.56 |
| | | *synPth* | 77.89 | 82.80 | 77.52 | 83.24 | 80.91 | 91.54 |
| J&N (89.28%) | Initial | *linPth* | 44.84 | 58.84 | 42.16 | 56.40 | 51.36 | 57.53 |
| | | *synPth* | 45.01 | 63.26 | 43.64 | 61.36 | 55.12 | 61.74 |
| | Selected | *linPth* | 77.73 | 83.21 | 76.45 | 82.70 | 80.15 | 89.77 |
| | | *synPth* | 77.70 | 83.90 | 76.79 | 83.71 | 80.88 | 90.59 |
| Gold (100%) | Initial | *linPth* | 45.57 | 61.79 | 42.41 | 59.09 | 53.12 | 53.12 |
| | | *synPth* | 45.89 | 67.63 | 43.76 | 65.51 | 57.77 | 57.77 |
| | Selected | *linPth* | 80.43 | 89.44 | 79.44 | 89.07 | 84.99 | 84.99 |
| | | *synPth* | 80.37 | 90.37 | 80.20 | 90.27 | 86.02 | 86.02 |

Table 6: Performance of random initial and greedily selected feature sets

the machine learning and linguistic perspective. The pruning provides a more balanced training dataset for classifier training than without pruning. Note that without pruning, most word pairs generated for the training are irrelevant and far away from the current predicate, inevitably interfering with the informative features from the truly relevant ones in the very small minority and, hence, leading to an unsatisfactory performance. Although the pruning, especially its adaptive version, is rooted in a linguistic insight gained from empirical observations on real data, most previous works on semantic parsing simply took the pruning as an indispensable step towards a good parsing performance, seldom paying much attention to the poor performance without pruning nor comparing it with the performance by different pruning strategies.

Table 6 presents a comprehensive results of our semantic dependency parsing on the three syntactic inputs aforementioned of different quality. A number of observations can be made from these results. (1) The greedy feature selection, as encoded in Algorithm 1 above, boosts the SRL performance drastically, raising the Sem-$F_1$ scores in the *synPth* rows from 54.79%–57.77% of the initial feature sets, the baseline, to 80.88%–86.02% of the selected feature sets, by an increment of 46.73%–48.90%. The rise in corresponding $linPth$ rows is even larger. Among the three inputs, the largest increment is on the gold standard, suggesting that the feature selection has a greater effect on an input of better quality. (2) The traverse scheme $synPth$ leads to a better model than $linPth$, as reflected in the difference of Sem-$F_1$ and Sem-$F_1$/LAS scores between them, indicating that this integrative SRL approach is sensitive to the path along which argument candidates are traversed. The difference of their Sem-$F_1$/LAS scores, for instance, is in the range of 7.14%–8.75% and 0.91%–1.21% for the initial and the selected feature sets, respectively. The significant advantage of $synPth$ is confirmed consistently, even though an optimized feature set narrows down the performance discrepancy between the two so radically. (3) The result that both Nomi-$F_{1N}^x$ and Verb-$F_{1P}^x$ are higher than corresponding $F_{1N+P}^x$ consistently throughout almost all experimental settings except one shows that the feature selection separately on Nombank or PropBank (for verbal or nominal predicates, respectively) gives a better performance than that on the combination Nombank+PropBank for both. This has to be explained by the interference between the two data sets due to their heterogeneous nature, namely, the interference between the nominal and verbal predicate samples. Hence, optimizing a feature set specifically for a particular type of predicates is more effective than for both. (4) An overall comparison of our system's SRL performance on the three syntactic inputs of different quality (as reflected in their LAS) shows that the performance as a whole varies in accord with the quality of input. This is exhibited in the contrast of the Sem-$F_1$ scores on these inputs, even though a small LAS difference may not necessarily lead to a significant performance difference (for instance, MST has a LAS of 0.89 percentage point lower than J&N but gives a Sem-$F_1$ score as high in one of the four experimental settings). The table also shows that a LAS difference of 11.61 percentage points, from 88.39% to 100%, corresponds to a Sem-$F_1$ score difference of at most 5.14 percentage points, from 80.88% to 86.02%, in the best setting (i.e., using the selected feature set and taking $synPth$).

| Syn. Input (LAS) | Feature set | Path $x$ | Nomi- $F_{1N}^x{}_{+P}$ | Verb- $F_{1N}^x{}_{+P}$ | Sem- $F_{1N}^x{}_{+P}$ | Sem- $F_{1N+P}^x$/LAS | Loss in $F_{1N+P}^x$ |
|---|---|---|---|---|---|---|---|
| MST (88.39%) | $S_{N+P}^I$ - Sense | $linPth$ | 76.51 | 82.09 | 79.82 | 90.30 | -0.29 |
| | $S_{N+P}^s$ - Sense | $synPth$ | 76.76 | 82.75 | 80.30 | 90.85 | -0.75 |
| | $S_N^I + S_P^I$ | $linPth$ | 76.78 | 82.20 | 79.99 | 90.50 | -0.07 |
| | $S_N^s + S_P^s$ | $synPth$ | 76.60 | 82.76 | 80.24 | 90.78 | -0.83 |

Table 7: Experimental results on feature ablation and feature set combination

However, Sem-$F_1$ scores cannot be trusted to faithfully reflect the competence of a semantic parser, because the quality of syntactic input is also a decisive factor to decide such scores. For this reason, we have the Sem-$F_1$/LAS ratio as an evaluation metric. Interestingly, our parser's scores of this ratio on the two syntactic inputs of a LAS 10.82–11.61 percentage points below the gold standard are, contrarily, 4.57–5.52 percentage points higher. This is certainly not to mean that the parser is able to rescue, in a sense, some true semantic parses from an erroneous syntactic input. Instead, it can only be explained by the parser's high tolerance of imperfections in the syntactic input.

Table 7 further presents experimental results on feature ablation and feature set combination. The former is to examine the effect of sense features and the latter that of feature optimization. Along *synPth*, both the ablation of sense feature and the mix of two feature sets respectively optimized (through the greedy selection) on the NomBank and PropBank lead to a significant performance loss of 0.75%–0.83%, in comparison with the performance of the feature set $S_{N+P}^s$ optimized on the combination of the two treebanks as given in Table 6. Along *linPth*, they lead to a much less significant and an insignificant loss, respestively. These results show that both sense features and the greedy selection of features are more significant in joining with the adaptive pruning along *synPth* to achieve a performance gain.

## 7.3 Comparison and Analysis

In order to evaluate the parser impartially in a comparative manner, its performance along *synPth* is compared with that of the other state-of-the-art systems in CoNLL-2008. They are chosen for this comparison because of being ranked among top four among all participants in the shared task or using some sophisticated joint learning techniques. The one of Titov, Henderson, Merlo, and Musillo (2009) that adopts a similar joint learning approach as Henderson, Merlo, Musillo, and Titov (2008) is also included, because of their significant methodological difference from the others. In particular, the former has attained the best performance to date in the direction of genuine joint learning. The reported performance of all these systems on the CoNLL-2008 test set in terms of a series of $F_1$ scores is presented in Table 8 for comparison. Ours is significantly better (t = 14.6, P < 0.025) than all the others except the post-evaluation result of Johansson and Nugues (2008). Contrary to the best three systems in CoNLL-2008 (Johansson & Nugues, 2008; Ciaramita, Attardi, Dell'Orletta, & Surdeanu, 2008; Che et al., 2008) that use SRL pipelines, our current work is intended to integrate them into one. Another baseline, namely, our current model using the feature set from the work of Zhao and Kit (2008), instead of a random set, is also included in the table for comparison, showing a significant performance enhancement on top of the previous model and, then, a further enhancement by the greedy feature selection.

Although this work draws necessary support from the basic techniques (especially those for traverse along *synPth*) underlying our previous systems for CoNLL-2008 and -2009 (Zhao & Kit, 2008; Zhao, Chen, Kit, & Zhou, 2009; Zhao, Chen, Kazama, Uchimoto, & Torisawa, 2009), what marks its uniqueness is that all SRL sub-tasks are performed by one integrative model with one selected feature set. Our previous systems dealt with predicate disambiguation as a separate sub-task. This is our first attempt at a fully integrated SRL system.

The fact that our integrated system is yet to give a performance on a par with the post-evaluation result of Johansson and Nugues (2008) seems attributable to a number of factors, including the ad hoc features adopted in their work to handle linguistic constructions such as raising/control and coordination. However, the most noticeable ones are the following discrepancies between the two systems, in addition to pipeline vs. all-in-one integration. (1) They have the n-best syntactic candidates as input, which without doubt provide more useful information than the 1-best that we use. (2) Then, they exploit reranking as a joint learning strategy to make fuller use of the n-best candidates and any intermediate semantic result once available, resulting in a gain of 0.5% increment of Sem-$F_1$ score. (3) They use respective sub-systems to deal with verbal and nominal predicates in a more specific manner, following the observation that adaptive optimization of feature sets for nominal

| Systems[a] | LAS | Sem-$F_1$ | Macro-$F_1$ | Sem-$F_1$/LAS | Pred-$F_1$[b] | Argu-$F_1$[c] | Verb-$F_1$ | Nomi-$F_1$ |
|---|---|---|---|---|---|---|---|---|
| Ours:Gold | 100.0 | 86.02 | 92.27 | 86.02 | 89.25 | 84.54 | 90.27 | 80.20 |
| Johansson:2008*[d] | 89.32 | 81.65 | 85.49 | 91.41 | 87.22 | 79.04 | 84.78 | 77.12 |
| Ours:MST | 88.39 | 80.91 | 85.09 | 91.54 | 87.15 | 78.01 | 83.23 | 77.52 |
| Ours:Johansson | 89.28 | 80.88 | 85.12 | 90.59 | 86.47 | 78.29 | 83.71 | 76.79 |
| Johansson:2008 | 89.32 | 80.37 | 84.86 | 89.98 | 85.40 | 78.02 | 84.45 | 74.32 |
| Ours:Baseline[e] | 88.39 | 79.42 | 84.34 | 89.85 | 86.60 | 76.08 | 81.71 | 76.07 |
| Ciaramita:2008* | 87.37 | 78.00 | 82.69 | 89.28 | 83.46 | 75.35 | 80.93 | 73.80 |
| Che:2008 | 86.75 | 78.52 | 82.66 | 90.51 | 85.31 | 75.27 | 80.46 | 75.18 |
| Zhao:2008* | 87.68 | 76.75 | 82.24 | 87.53 | 78.52 | 75.93 | 78.81 | 73.59 |
| Ciaramita:2008 | 86.60 | 77.50 | 82.06 | 89.49 | 83.46 | 74.56 | 80.15 | 73.17 |
| Titov:2009 | 87.50 | 76.10 | 81.80 | 86.97 | – | – | – | – |
| Zhao:2008 | 86.66 | 76.16 | 81.44 | 87.88 | 78.26 | 75.18 | 77.67 | 73.28 |
| Henderson:2008* | 87.64 | 73.09 | 80.48 | 83.40 | 81.42 | 69.10 | 75.84 | 68.90 |
| Henderson:2008 | 86.91 | 70.97 | 79.11 | 81.66 | 79.60 | 66.83 | 73.80 | 66.26 |

a. Ranked according to Sem-$F_1$, and only first authors are listed for the sake of space limitation.

b. Labeled $F_1$ for predicate identification and classification.

c. Labeled $F_1$ for argument identification and classification.

d. A superscript * indicates post-evaluation results, available from the official website of CoNLL-2008 shared task at http://www.yr-bcn.es/dokuwiki/doku.php?id=conll2008:start.

e. Syntactic input and traverse scheme: as Ours:MST; Features: as Zhao:2008

Table 8: Performance comparison of the best existing SRL systems

or verbal predicates respectively is more likely to give a better performance than that for a mix of both. This observation is also confirmed by evidence in our experimental results: $F1_N^x$ and $F1_P^x$ scores are consistently higher than respective $F1_{N+P}^x$ ones in Table 6 above.

Because of the integrative nature of our approach, however, our priority has to be given to optimizing the whole feature set for both verbal and nominal predicates. It is nevertheless understood that all these point to potential ways to further enhance our system, e.g., by taking advantage of specialized feature sets for various kinds of words and/or utilizing some joint learning techniques such as syntactic-semantic reranking, in a way that the integrity of the system can be maintained properly.

The difference between the joint learning in the work of Johansson and Nugues (2008) and that of Titov et al. (2009) is worth noting. The former is a kind of cascade-style joint learning that first has a syntactic submodel to provide the n-best syntactic trees and a semantic submodel to infer correspondent semantic structures, and then a reranking model, with the log probabilities of the syntactic trees and semantic structures as its features, to find the best joint syntactic-semantic analysis, resulting in an improvement on top of individual submodels. In contrast to the former with a non-synchronous pipeline from syntactic to semantic parsing, the latter adopts a stricter all-in-one strategy of joint learning, where syntactic and semantic dependencies are learnt and decoded synchronously, based on an augmented version of the transition-based shift-reduce parsing strategy (Henderson et al., 2008). Regrettably, however, the performance of this approach is still far from the top of the ranked list in Table 8, indicating the particular significance of our current work.

Whether it is worth integrating some form of joint-learning into an integrative system such as ours depends on the cost-effectiveness of doing so. It has been illustrated that such joint learning does lead to certain performance improvement, as in CoNLL shared task on SRL and successive works, e.g., by Johansson and Nugues (2008). However, a great deal of computational cost has to be paid in order to enable such a reranking procedure to handle multiple syntactic inputs. This certainly makes it impractical for real applications, not to mention that an integrative system is born with a particularly strong demand for integrity to preclude itself from accommodating such a stand-alone submodel.

## 8. Conclusion

Semantic parsing, which aims to derive and instantiate the semantic structure of a sentence via identifying semantic relations between words, plays a critical role in deep processing of natural language. In this article, we have presented an integrative approach to semantic dependency parsing in the form of semantic role labeling, its implementation as an all-in-one word pair classifier, and a comprehensive evaluation of it using three syntactic inputs of different quality. The evaluation results confirm the effectiveness and practicality of this approach. The major contributions of this research are the following. It exhibits a significant success for the first time that an integrative SRL system has achieved a performance next only to that of the best pipeline system, indicating the potentials of the integrative approach besides its practicality for real applications. The large-scale feature selection engineering underlying the success of this work also demonstrates (1) how the largest feature space ever in use in this field is formed by allowing a wide range of flexible (re)combinations of basic elements extracted from the known features and properties of input words and (2) how a speedy adaptive feature selection procedure is formulated and applied to select the most effective set of features from the allowable feature space.

The core techniques that have contributed to this success are developed based on the two types of traverse path, along syntactic tree branches vs. linear input word sequence. Both argument candidate pruning and feature selection are performed along an identical path. The strategy of using auxiliary labels to facilitate argument candidate pruning, following the observation that true arguments tend to be close to their predicates, works well with both traverse schemes. Interestingly, although the feature selection procedure outputs two very different feature sets for each of NomBank, PropBank and their combination whilst working along the two paths, both feature sets lead the SRL system to a very close performance on the same test data, a competitive performance on top of all but one best pipeline system, confirming the robustness and effectiveness of the feature selection procedure.

Evidence is also presented in our evaluation results to reconfirm the finding in the previous works of semantic parsing that feature sets optimized specifically for verbal or nominal predicates outperform a collective one for both. However, the competitive performance of the collective one that we have arrived at also suggests that a harmonious rival feature set for both types of predicate as a whole is reachable and its slight performance difference from the specific sets is fairly acceptable as the unavoidable small cost for exchange for the higher integrity and practicality of an integrative SRL system. This competitiveness is attributable at least to two main factors. One is the very large feature space in use, which provides about a dozen times as many feature templates as those in the previous works (e.g., see Xue & Palmer, 2004; Xue, 2006). The other is the ME classifier that can accommodate so many features in one model. According to our experience in this piece of work, the ME model is not vulnerable to the use of many overlapping features, from which SVM and other margin-based learners usually suffer a lot.

## Acknowledgments

The research reported in this article was partially supported by the Department of Chinese, Translation and Linguistics, City University of Hong Kong, through a post-doctorate research fellowship to the first author and a research grant (CTL UNFD-GRF-144611) to the third and corresponding author, the National Natural Science Foundation of China (Grants 60903119 and 61170114), the National Basic Research Program of China (Grant 2009CB320901), the National High-Tech Research Program of China (Grant 2008AA02Z315), the Research Grants Council of HKSAR, China (Grant CityU 144410), and the City University of Hong Kong (Grant 7002796). Special thanks are owed to Richard Johansson for kindly providing his syntactic output for the CoNLL-2008 shared task, to three anonymous reviewers for their insightful comments and to John S. Y. Lee for his help.

## Appendix A. Feature Templates and their Importance Rankings

| Type | PropBank | | | | Extra/Auxiliary | Total |
|---|---|---|---|---|---|---|
| Predicate | 01˜21 (21) | | | | NONE_PRED | 22 |
| Argument | A0˜5 | AM-ADV | C-AM-ADV | R-AM-ADV | NONE_ARG | |
| | AA, AM | AM-CAU | C-AM-CAU | R-AM-CAU | | |
| | C-A0˜4 | AM-DIR | C-AM-DIR | R-AM-DIR | noMoreArg | |
| | R-A0˜4 | AM-DIS | C-AM-DIS | R-AM-EXT | (for *synPth*) | 56 |
| | R-AA | AM-EXT | C-AM-EXT | R-AM-LOC | | |
| | AM-PRD | AM-LOC | C-AM-LOC | R-AM-MNR | noMoreLeftArg | |
| | AM-PRT | AM-MNR | C-AM-MNR | R-AM-PNC | noMoreRighArg | |
| | AM-REC | AM-MOD | C-AM-NEG | R-AM-TMP | (for *linPth*) | 57 |
| | AM-TM | AM-NEG | C-AM-PNC | C-R-AM-TMP | | |
| | AM-TMP | AM-PNC | C-AM-TMP | SU (54) | | |

Table 9: The list of class labels for predicate and argument

| | Template | Rank in: | $S^s_{N+P}$ | $S^s_N$ | $S^s_P$ | $S^l_{N+P}$ | $S^l_N$ | $S^l_P$ |
|---|---|---|---|---|---|---|---|---|
| - | p.lm.dprel | | 41 | 39 | 6 | 82 | 113 | 60 |
| - | a:p\|dpPath.dprel | | 35 | 31 | 52 | 2 | 62 | 2 |
| - | a.lemma + p.lemma | | 10 | 44 | 4 | 5 | 36 | 6 |
| - | a.lemma + a.dprel + a.h.lemma | | 55 | 40 | 49 | 112 | 69 | 44 |
| - | a.spLemma + p.spLemma | | 4 | 97 | 15 | 13 | 68 | 26 |

Table 10: Overlap of the six resulted feature template sets

| Template | Rank in: | $S_{N+P}^s$ | $S_N^s$ | $S_P^s$ |
|---|---|---|---|---|
| p$_{-1}$.pos + p.pos | | 2 | 37 | 79 |
| p$_{-1}$.spLemma | | 27 | 13 | 59 |
| p.spForm + p.lm.spPos + p.noFarChildren.spPos.bag + p.rm.spPos | | 7 | 45 | 63 |
| a.isCurPred.lemma | | 83 | 94 | 75 |
| a.isCurPred.spLemma | | 36 | 38 | 86 |
| a:p\|existCross | | 48 | 77 | 82 |
| a:p\|dpPath.dprel.bag | | 47 | 14 | 85 |
| a:p\|dpPathPred.spForm.bag | | 97 | 24 | 5 |
| a:p\|dpPath.spLemma.seq | | 67 | 59 | 71 |
| a:p\|linePath.spForm.bag | | 85 | 48 | 61 |
| a.semdprel = A$_0$ ? | | 50 | 86 | 40 |

Table 11: Overlap of $S_N^s$, $S_P^s$ and $S_{N+P}^s$ besides Table 10

| Template | Rank in: | $S_{N+P}^l$ | $S_N^l$ | $S_P^l$ |
|---|---|---|---|---|
| p.spLemma + p.currentSense | | 18 | 28 | 56 |
| p.currentSense + a.spLemma | | 33 | 57 | 17 |
| p.voice + (a:p\|direction) | | 65 | 120 | 25 |
| p.children.dprel.noDup | | 11 | 54 | 40 |
| p.rm.dprel | | 60 | 114 | 3 |
| p.rm.form | | 113 | 110 | 80 |
| p$_{-1}$.spLemma + p.spLemma | | 38 | 61 | 69 |
| p.voice | | 26 | 4 | 10 |
| p.form + p.children.dprel.noDup | | 96 | 81 | 65 |
| p.lm.form + p.noFarChildren.spPos.bag + p.rm.form | | 88 | 106 | 5 |
| p.lemma | | 4 | 26 | 50 |
| p.lemma + p$_1$.lemma | | 7 | 5 | 34 |
| p.spForm | | 39 | 100 | 36 |
| p.spForm + p.children.dprel.bag | | 91 | 6 | 30 |
| p.spForm + p.lm.spForm + p.noFarChildren.spPos.bag + p.rm.spForm | | 104 | 10 | 14 |
| p.splemma | | 9 | 65 | 64 |
| p.spLemma + p.h.spForm | | 100 | 11 | 70 |
| p.spLemma + p$_1$.spLemma | | 72 | 112 | 33 |
| p$_1$.pos | | 76 | 104 | 28 |
| a$_{-1}$.isCurPred.lemma | | 67 | 109 | 77 |
| a$_{-1}$.isCurPred.lemma + a.isCurPred.lemma | | 42 | 24 | 43 |
| a$_{-1}$.isCurPred.spLemma + a.isCurPred.spLemma | | 14 | 89 | 62 |
| a.isCurPred.Lemma + a$_1$.isCurPred.Lemma | | 29 | 44 | 67 |
| a.isCurPred.spLemma + a$_1$.isCurPred.spLemma | | 50 | 45 | 59 |
| a.spPos.baseline_Ax + a.voice + (a:p\|direction) | | 24 | 9 | 46 |
| a.spPos.baseline_Mod | | 86 | 80 | 18 |
| a.h.children.dprel.bag | | 97 | 35 | 45 |
| a.lm$_{-1}$.spPos | | 47 | 63 | 49 |
| a.lm$_1$.lemma | | 49 | 30 | 68 |
| a.children.spPos.seq + p.children.spPos.seq | | 19 | 90 | 76 |
| a.rm.dprel + a.pos | | 21 | 17 | 24 |
| a.rm.dprel + a.spPos | | 30 | 7 | 22 |
| a.rm$_{-1}$.spPos | | 6 | 74 | 15 |
| a.rm.lemma | | 36 | 50 | 4 |

| Template | | | |
|---|--:|--:|--:|
| a.rn.dprel + a.spPos | 28 | 33 | 72 |
| $a_{-1}$.lemma + a.lemma | 27 | 46 | 37 |
| a:p\|dpPathArgu.dprel.seq | 3 | 96 | 1 |
| a:p\|dpPathArgu.pos.seq | 75 | 79 | 9 |
| a:p\|dpPathPred.dprel.seq | 12 | 64 | 35 |
| a.form | 53 | 94 | 78 |
| a.form + a.pos | 32 | 93 | 32 |
| a.form + $a_1$.form | 94 | 31 | 38 |
| a.spForm + a.spPos | 16 | 73 | 48 |
| a.spForm + $a_1$.spForm | 79 | 38 | 52 |
| a.spLemma + a.dprel | 43 | 118 | 7 |
| a.spLemma + a.h.spForm | 110 | 2 | 51 |

Table 12: Overlap of $S_N^l$, $S_P^l$ and $S_{N+P}^l$ besides Table 10

| Template | Rank | Template | Rank |
|---|--:|---|--:|
| p.lemma + p.currentSense | 82 | p.spLemma + p.currentSense | 80 |
| p.currentSense + a.lemma | 57 | p.currentSense + a.spLemma | 18 |
| a.form + p.semdprel is ctype ? | 3 | a.form + p.ctypeSemdprel | 4 |
| a.form + p.semdprel is rtype ? | 5 | a.form + p.rtypeSemdprel | 6 |
| p.lm.form | 47 | p.lm.spForm | 7 |
| $p_{-1}$.form + p.form | 71 | $p_{-1}$.spLemma + p.spLemma | 92 |
| $p_{-2}$.form | 78 | $p_{-2}$.spForm | 61 |
| $p_{-2}$.spForm + $p_{-1}$.spForm | 15 | p.form | 68 |
| p.form + p.dprel | 74 | p.lemma | 63 |
| p.lemma + p.h.form | 10 | p.pos | 62 |
| p.spForm + p.dprel | 46 | p.spForm + p.children.dprel.bag | 90 |
| p.spLemma + p.children.dprel.noDup | | p.spLemma + p.h.spForm | 27 |
| p.spLemma + $p_1$.spLemma | 49 | $p_1$.pos | 28 |
| a.voice + (a:p\|direction) | 23 | a.children.adv.bag | 95 |
| a is leaf in syntactic tree ? | 16 | a.lm.dprel + a.form | 75 |
| a.lm.dprel + a.spPos | 67 | $a.lm_{-1}$.spLemma | 100 |
| a.lm.pos + a.pos | 50 | a.lm.spPos | 8 |
| a.pphead.spLemma | 19 | a.rm.dprel + a.spPos | 26 |
| $a.rm_{-1}$.form | 81 | $a.rm_{-1}$.spForm | 55 |
| $a.rm_1$.spPos | 79 | a.rn.dprel + a.spForm | 32 |
| a.highSupportVerb.form | 56 | a.highSupportVerb.spForm | 99 |
| a.lowSupportPorp.form | 51 | a.lowSupportPorp.lemma | 91 |
| a.lowSupportPorp.spLemma | 69 | $a_{-1}$.lemma + $a_1$.lemma | 20 |
| $a_{-1}$.pos | 70 | $a_{-1}$.pos + a.pos | 84 |
| $a_{-1}$.spForm | 85 | $a_{-1}$.spPos + $a_1$.spPos | 98 |
| a:p\|dpPath.distance | 9 | a:p\|dpPath.spLemma.bag | 73 |
| a:p\|dpPathArgu.spLemma.bag | 96 | a:p\|dpPathPred.spLemma.bag | 2 |
| a:p\|dpPathPred.spPos.bag | 93 | a:p\|dpPathArgu.dprel.seq | 22 |
| a:p\|linePath.dprel.bag | 88 | a.semdprel = $A_2$ ? | 35 |
| a.form + a.children.pos.seq | 53 | a.form + a.form | 58 |
| a.form + a.pos | 1 | a.pos + a.children.spPos.seq | 12 |
| a.spForm + a.children.spPos.seq | 76 | a.spForm + a.children.spPos.bag | 65 |
| a.spForm + a.spPos | 87 | a.spForm + $a_1$.spForm | 52 |
| a.spLemma | 11 | a.spLemma + a.pphead.spForm | 64 |
| a.spLemma + $a_1$.spLemma | 60 | a.spPos + a.dprel + a.h.spPos | 41 |

| Template | Rank | Template | Rank |
|---|---|---|---|
| a$_1$.form | 54 | a$_1$.spForm | 83 |
| a$_1$.spPos | 33 | (a:p|dpTreeRelation) + p.form | 25 |
| (a:p|dpTreeRelation) + p.spPos | 29 | (a:p|dpTreeRelation) + a.spPos | 30 |
| (a:p|dpPath.dprel.seq) + p.spForm | 36 | | |
| a$_{-1}$.isCurPred.spLemma + a.isCurPred.spLemma | 17 | | |
| a.noFarChildren.spPos.bag + a.rm.spPos | 21 | | |
| a.children.spPos.seq + p.children.spPos.seq | 34 | | |
| a.highSupportNoun:p|dpPath.dprel.seq | 89 | | |
| (a.highSupportNoun:p|dpTreeRelation) + p.form | 66 | | |
| (a.highSupportVerb:p|dpTreeRelation) + a.spPos | 72 | | |
| (a.lowSupportVerb:p|dpTreeRelation) + a.spForm | 42 | | |

Table 13: Feature templates of $S_N^x$ besides Tables 10 and 11

| Template | Rank | Template | Rank |
|---|---|---|---|
| p.rm.dprel | 47 | p.dprel | 25 |
| p.children.dprel.bag | 66 | p.lm.spPos | 48 |
| p.children.pos.seq | 70 | p.rm.dprel | 51 |
| p$_{-2}$.pos | 23 | p$_{-2}$.spForm + p$_{-1}$.spForm | 43 |
| p.dprel = OBJ ? | 50 | p.lemma + p.h.form | 68 |
| p.lemma+p$_1$.lemma | 3 | p.pos | 26 |
| p.spForm | 76 | p.spForm + p.children.dprel.noDup | 60 |
| p.splemma | 9 | p.spLemma+p$_1$.spLemma | 1 |
| p$_1$.spPos | 21 | a.lowSupportVerb:p|dpTreeRelation | 32 |
| a.children.adv.bag | 20 | a.dprel | 13 |
| a.children.dprel.bag | 7 | a.h.lemma | 8 |
| a.h.spLemma | 72 | a.lm.dprel + a.spPos | 31 |
| a.lm$_{-1}$.spLemma | 54 | a.pphead.lemma | 80 |
| a.pphead.spLemma | 46 | a.rm.dprel + a.spPos | 78 |
| a$_{-1}$.lemma + a$_1$.lemma | 16 | a$_{-1}$.pos | 24 |
| a$_{-1}$.spLemma + a.spLemma | 29 | a:p|linePath.distance | 55 |
| a:p|dpPath.distance | 22 | a:p|dpPathPred.dprel.bag | 53 |
| a:p|dpPath.spForm.seq | 12 | a:p|dpPathArgu.spForm.seq | 11 |
| a:p|dpPathArgu.spLemma.bag | 84 | a:p|dpPathPred.spLemma.bag | 2 |
| a:p|dpPathArgu.spLemma.seq | 17 | a:p|dpPath.spPos.bag | 65 |
| a:p|dpPathPred.spPos.bag | 64 | a:p|dpPathArgu.dprel.seq | 28 |
| a:p|linePath.spLemma.seq | 42 | a:p|linePath.spLemma.bag | 27 |
| a:p|dpPathPred.spPos | 62 | a.existSemdprel.A$_0$ | 67 |
| a.existSemdprel.A$_1$ | 56 | a.existSemdprel.A$_2$ | 57 |
| a.semdprel = A$_2$ ? | 77 | a.dprel = OBJ ? | 73 |
| a.form + a.children.pos.seq | 10 | a.pos + p.pos | 45 |
| a.spLemma + a.dprel | 87 | a.spLemma+a.dprel+a.h.spLemma | 19 |
| a$_1$.lemma | 14 | a$_1$.spPos | 69 |
| (a:p|dpTreeRelation) + a.spPos | 81 | (a:p|dpPath.dprel.seq) + p.spPos | 18 |
| a$_{-1}$.isCurPred.spLemma + a.isCurPred.spLemma | 41 | | |
| a$_{-2}$.isCurPred.lemma + a$_{-1}$.isCurPred.lemma | 58 | | |
| a.isCurPred.spLemma + a$_1$.isCurPred.spLemma | 74 | | |
| a.lowSupportVerb:p|dpPath.dprel.seq | 33 | | |
| a.lowSupportVerb:p|dpPathArgu.dprel.seq | 34 | | |
| a.lowSupportVerb:p|dpPathArgu.spPos.seq | 35 | | |
| a.lowSupportVerb:p|dpPathShared.dprel.seq | 36 | | |

| | |
|---|---|
| `a.lowSupportVerb:p\|dpPathShared.spPos.seq` | 37 |
| `a.lowSupportVerb:p\|dpPathPred.dprel.seq` | 38 |
| `a.lowSupportVerb:p\|dpPathPred.spPos.seq` | 39 |
| `a.highSupportNoun:p\|dpPath.dprel.seq` | 83 |
| `a.lowSupportVerb:p\|dpPath.dprel.seq` | 30 |
| `(a.highSupportVerb:p\|dpTreeRelation) + a.spPos` | 44 |

Table 14: Feature templates of $S_P^s$ besides Tables 10 and 11

| Template | Rank | Template | Rank |
|---|---|---|---|
| `p.lemma + p.currentSense` | 100 | `p.currentSense + a.lemma` | 61 |
| `a.form + p.semdprel is ctype ?` | 90 | `a.form + p.ctypeSemdprel` | 91 |
| `a.form + p.semdprel is rtype ?` | 92 | `a.form + p.rtypeSemdprel` | 93 |
| `p.dprel` | 8 | `p.children.pos.seq` | 6 |
| `p.rm.dprel` | 46 | `p.lowSupportProp:p\|dpTreeRelation` | 12 |
| `p_{-1}.spForm + p.spForm` | 54 | `p.voice` | 9 |
| `p.lemma+p_1.lemma` | 18 | `p.pos + p.dprel` | 5 |
| `p.splemma` | 88 | `p.spLemma+p.h.spForm` | 3 |
| `p.spPos + p.children.dprel.bag` | 15 | `p.spPos + p_1.spPos` | 14 |
| `p_1.spForm` | 26 | `p_{-1}.isCurPred.lemma` | 28 |
| `a.isCurPred.pos` | 84 | `a.isCurPred.spPos` | 96 |
| `a_1.isCurPred.Lemma` | 37 | `a_1.isCurPred.spLemma` | 22 |
| `a:p\|direction` | 57 | `(a:p\|dpPath.dprel.seq) + a.spForm` | 11 |
| `a.form.baseline_Mod` | 73 | `a.pos.baseline_Mod` | 74 |
| `a.spForm.baseline_Mod` | 75 | `a.baseline_Mod` | 76 |
| `a.lm.Lemma` | 59 | `a.lm.spForm` | 60 |
| `a.lm.spPos` | 65 | `a.rm.lemma` | 81 |
| `a.highSupportNoun.pos` | 62 | `a.highSupportNoun.spPos` | 20 |
| `a.highSupportVerb.spPos` | 42 | `a.lowSupportNoun.pos` | 87 |
| `a.lowSupportPorp.spLemma` | 98 | `a.lowSupportVerb.pos` | 56 |
| `a_{-1}.spLemma+a.spLemma` | 38 | `a:p\|dpPathPred.spLemma.seq` | 63 |
| `a:p\|linePath.spForm.seq` | 80 | `a:p\|linePath.spLemma.seq` | 53 |
| `a:p\|linePath.spLemma.bag` | 86 | `a:p\|linePath.spPos.seq` | 51 |
| `a:p\|linePath.spPos.bag` | 66 | `a:p\|dpPathPred.spPos` | 39 |
| `a.existSemdprel_A_0` | 49 | `a.existSemdprel_A_1` | 1 |
| `a.form` | 94 | `a.form = p.form ?` | 40 |
| `a.form + a.form` | 95 | `a.lemma` | 43 |
| `a.lemma + a.dprel` | 21 | `a.lemma + a.h.form` | 29 |
| `a.lemma + a.pphead.form` | 44 | `a.spForm = p.spForm ?` | 89 |
| `a.spLemma + a.pphead.spForm` | 24 | `a.spPos + a.spPos` | 82 |
| `(a:p\|dpPath.dprel.seq) + p.form` | 45 | `(a:p\|dpPath.dprel.seq) + p.spForm` | 25 |
| `(a:p\|dpPath.dprel.seq) + a.form` | 13 | | |
| `p.lm.form + p.noFarChildren.spPos.bag + p.rm.form` | | | 52 |
| `a_{-2}.isCurPred.lemma + a_{-1}.isCurPred.lemma` | | | 64 |
| `a.isCurPred.pos + a_1.isCurPred.pos` | | | 99 |
| `a.isCurPred.spLemma + a_1.isCurPred.spLemma` | | | 23 |
| `a.form.baseline_Ax + a.voice + (a:p\|direction)` | | | 77 |
| `a.spForm.baseline_Ax+ a.voice + (a:p\|direction)` | | | 78 |
| `a.spPos.baseline_Ax + a.voice + (a:p\|direction)` | | | 79 |
| `a.highSupportNoun:p\|dpPathShared.dprel.seq` | | | 30 |
| `a.highSupportVerb:p\|dpPathShared.dprel.seq` | | | 68 |

| Template | Rank |
|---|---|
| ⌐ a.lowSupportNoun:p\|dpPath.dprel.seq | 16 |
| ⌐ a.lowSupportNoun:p\|dpPathArgu.dprel.seq | 31 |
| ⌐ a.lowSupportNoun:p\|dpPathArgu.spPos.seq | 32 |
| ⌐ a.lowSupportNoun:p\|dpPathShared.dprel.seq | 33 |
| ⌐ a.lowSupportNoun:p\|dpPathShared.spPos.seq | 34 |
| ⌐ a.lowSupportNoun:p\|dpPathPred.dprel.seq | 17 |
| ⌐ a.lowSupportVerb:p\|dpPathArgu.dprel.seq | 69 |
| ⌐ a.lowSupportVerb:p\|dpPathArgu.spPos.seq | 70 |
| ⌐ a.lowSupportVerb:p\|dpPathShared.dprel.seq | 71 |
| ⌐ a.lowSupportVerb:p\|dpPathShared.spPos.seq | 72 |
| ⌐ (a.highSupportVerb:p\|dpTreeRelation) + a.form | 58 |
| ⌐ (a.lowSupportNoun:p\|dpTreeRelation) + p.spPos | 19 |

Table 15: Feature templates of $S_{N+P}^s$ besides Tables 10 and 11

| Template | Rank | Template | Rank |
|---|---|---|---|
| ⌐ p$_{-1}$.spLemma | 74 | ⌐ p$_{-2}$.form | 55 |
| ⌐ p$_1$.spPos | 19 | ⌐ a$_1$.isCurPred.Lemma | 71 |
| ⌐ a$_1$.isCurPred.spLemma | 53 | ⌐ a.children.dprel.bag | 42 |
| ⌐ a.h.lemma | 23 | ⌐ a.lm.dprel + a.pos | 63 |
| ⌐ a.lm$_{-1}$.lemma | 31 | ⌐ a.lm.Lemma | 29 |
| ⌐ a.pphead.lemma | 27 | ⌐ a.pphead.spLemma | 39 |
| ⌐ a.lowSupportNoun.spPos | 8 | ⌐ a.lowSupportPorp.form | 73 |
| ⌐ a.lowSupportPorp.lemma | 47 | ⌐ a.lowSupportPorp.spForm | 79 |
| ⌐ a.lowSupportPorp.spLemma | 57 | ⌐ a$_{-1}$.spPos | 58 |
| ⌐ a$_{-1}$.spPos + a$_1$.spPos | 54 | ⌐ a.semdprel = A$_2$ ? | 20 |
| ⌐ (a:p\|dpTreeRelation) + p.pos | 41 | ⌐ (a:p\|dpTreeRelation) + p.spPos | 21 |
| ⌐ a$_{-2}$.isCurPred.spLemma + a$_{-1}$.isCurPred.spLemma | 61 | | |
| ⌐ a.lowSupportPorp:p\|dpPathShared.dprel.seq | 12 | | |
| ⌐ a.lowSupportPorp:p\|dpPathShared.spPos.seq | 13 | | |
| ⌐ a.lowSupportVerb:p\|dpPath.dprel.seq | 16 | | |
| ⌐ (a.highSupportVerb:p\|dpTreeRelation) + a.form | 11 | | |
| ⌐ (a.lowSupportNoun:p\|dpTreeRelation) + p.pos | 75 | | |
| ⌐ (a.lowSupportNoun:p\|dpTreeRelation) + p.spPos | 66 | | |

Table 16: Feature templates of $S_P^l$ besides Tables 10 and 12

| Template | Rank | Template | Rank |
|---|---|---|---|
| ⌐ p.rm.dprel | 88 | ⌐ p.children.dprel.seq | 27 |
| ⌐ p.lowSupportNoun.spForm | 16 | ⌐ p.lowSupportProp:p\|dpTreeRelation | 72 |
| ⌐ p$_{-1}$.form + p.form | 103 | ⌐ p$_{-1}$.lemma + p.lemma | 91 |
| ⌐ p$_{-1}$.pos+p.pos | 32 | ⌐ p$_{-1}$.spForm + p.spForm | 40 |
| ⌐ p$_{-1}$.spLemma | 13 | ⌐ p$_{-2}$.form + p$_{-1}$.form | 99 |
| ⌐ p$_{-2}$.pos | 18 | ⌐ p$_{-2}$.spForm | 39 |
| ⌐ p.dprel = OBJ ? | 59 | ⌐ p.form + p.dprel | 95 |
| ⌐ p.lemma + p.h.form | 42 | ⌐ p.pos + p.dprel | 1 |
| ⌐ p.spPos + p$_1$.spPos | 34 | ⌐ p$_1$.spForm | 86 |
| ⌐ a.voice + (a:p\|direction) | 75 | ⌐ a.isCurPred.lemma | 43 |
| ⌐ a.isCurPred.spLemma | 29 | ⌐ a.lm.dprel + a.dprel | 98 |

| Template | Rank | Template | Rank |
|---|---|---|---|
| a.lm.dprel + a.pos | 76 | a.lm$_{-1}$.spLemma | 3 |
| a.lm.pos + a.pos | 19 | a.lm.spForm | 107 |
| a.lm.spPos | 49 | a.ln.dprel + a.pos | 25 |
| a.rm$_1$.spPos | 21 | a.highSupportNoun.lemma | 14 |
| a.highSupportNoun.pos | 48 | a.highSupportNoun.spPos | 51 |
| a.lowSupportVerb.pos | 97 | a.lowSupportVerb.spLemma | 78 |
| a.lowSupportVerb.spPos | 12 | a$_{-1}$.lemma | 101 |
| a$_{-1}$.spLemma+a.spLemma | 77 | a$_{-2}$.pos | 102 |
| a:p\|linePath.distance | 67 | a:p\|dpTreeRelation | 20 |
| a:p\|dpPathPred.spPos | 115 | a.dprel = OBJ ? | 116 |
| a.form + p.form | 83 | a.pos + p.pos | 92 |
| a.spForm + p.spForm | 87 | a.spForm + a.children.spPos.seq | 53 |
| a.spForm + a.children.spPos.bag | 119 | a.spLemma+a.dprel+a.h.spLemma | 60 |
| a.spLemma + a.pphead.spForm | 66 | a.spLemma + a$_1$.spLemma | 55 |
| a$_1$.pos | 52 | a$_1$.spPos | 23 |
| (a:p\|dpTreeRelation) + p.form | 111 | (a:p\|dpTreeRelation) + p.spForm | 8 |
| (a:p\|dpTreeRelation) + a.form | 84 | (a:p\|dpTreeRelation) + a.spForm | 41 |
| (a:p\|dpTreeRelation) + a.spPos | 15 | (a:p\|dpPath.dprel.seq) + p.form | 56 |
| (a:p\|dpPath.dprel.seq) + p.spForm | 108 | (a:p\|dpPath.dprel.seq) + a.form | 70 |
| (a:p\|dpPath.dprel.seq) + a.spForm | 37 | | |
| p.lm.spPos + p.noFarChildren.spPos.bag + p.rm.spPos | | | 117 |
| a$_{-2}$.isCurPred.lemma + a$_{-1}$.isCurPred.lemma | | | 58 |
| (a$_1$:p\|direction) + (a$_2$:p\|direction) | | | 105 |
| a.noFarChildren.spPos.bag + a.rm.spPos | | | 22 |
| a.highSupportVerb:p\|dpTreeRelation | | | 85 |
| (a.highSupportVerb:p\|dpTreeRelation) + a.form | | | 47 |
| (a.lowSupportNoun:p\|dpTreeRelation) + p.form | | | 82 |
| (a.lowSupportNoun:p\|dpTreeRelation) + p.spForm | | | 71 |

Table 17: Feature templates of $S_N^l$ besides Tables 10 and 12

| Template | Rank | Template | Rank |
|---|---|---|---|
| p.currentSense + a.spPos | 69 | p.rm.dprel | 117 |
| p.lm.form | 101 | p.lm.spForm | 51 |
| p.lowSupportNoun.spForm | 99 | p.lowSupportProp:p\|dpTreeRelation | 74 |
| p$_{-1}$.form + p.form | 106 | p$_{-1}$.pos+p.pos | 1 |
| p$_{-1}$.spForm + p.spForm | 98 | p$_{-2}$.form + p$_{-1}$.form | 40 |
| p$_{-2}$.pos | 87 | p$_{-2}$.spForm | 54 |
| p.form + p.dprel | 114 | p.spForm + p.dprel | 115 |
| p.spPos + p$_1$.spPos | 45 | p$_1$.spForm | 37 |
| p$_1$.spPos | 102 | a.voice + (a:p\|direction) | 10 |
| a.isCurPred.lemma | 52 | a.isCurPred.spLemma | 66 |
| a$_1$.isCurPred.Lemma | 41 | a$_1$.isCurPred.spLemma | 64 |
| a.children.dprel.bag | 48 | a.lm.dprel + a.dprel | 70 |
| a.lm$_{-1}$.lemma | 20 | a.lm$_{-1}$.spLemma | 17 |
| a.lm.Lemma | 84 | a.lm.pos + a.pos | 8 |
| a.lm.spForm | 34 | a.lm.spPos | 59 |
| a.ln.dprel + a.pos | 63 | a.rm$_1$.spPos | 111 |
| a.lowSupportNoun:p\|dpTreeRelation | 93 | a.lowSupportVerb.spLemma | 15 |
| a$_{-1}$.lemma | 81 | a$_{-1}$.spLemma+a.spLemma | 31 |
| a$_{-1}$.spPos | 109 | a$_{-1}$.spPos + a$_1$.spPos | 92 |

| | | | | | |
|---|---|---|---|---|---|
| _ | `a:p\|linePath.distance` | 80 | _ | `a:p\|dpTreeRelation` | 57 |
| _ | `a:p\|dpPathPred.spPos` | 85 | _ | `a.existSemdprel_A_2` | 77 |
| _ | `a.semdprel = A_2 ?` | 78 | _ | `a.spForm + a.children.spPos.seq` | 71 |
| _ | `a.spForm + a.children.spPos.bag` | 61 | _ | `a.spLemma+a.dprel+a.h.spLemma` | 90 |
| _ | `a.spLemma + a.pphead.spForm` | 62 | _ | `a_1.lemma` | 68 |
| _ | `a_1.spPos` | 44 | _ | `(a:p\|dpTreeRelation) + a.form` | 25 |
| _ | `(a:p\|dpTreeRelation) + a.spForm` | 73 | _ | `(a:p\|dpTreeRelation) + a.spPos` | 58 |
| _ | `(a:p\|dpPath.dprel.seq) + p.form` | 22 | _ | `(a:p\|dpPath.dprel.seq) + p.spForm` | 83 |
| _ | `(a:p\|dpPath.dprel.seq) + a.form` | 89 | _ | `(a:p\|dpPath.dprel.seq) + a.spForm` | 103 |
| _ | `p.spForm + p.lm.spPos + p.noFarChildren.spPos.bag + p.rm.spPos` | 108 | | | |
| _ | `a_{-2}.isCurPred.lemma + a_{-1}.isCurPred.lemma` | 23 | | | |
| _ | `a_{-2}.isCurPred.spLemma + a_{-1}.isCurPred.spLemma` | 46 | | | |
| _ | `a.noFarChildren.spPos.bag + a.rm.spPos` | 95 | | | |
| _ | `a.highSupportNoun:p\|dpPath.dprel.seq` | 55 | | | |
| _ | `a.lowSupportVerb:p\|dpPath.dprel.seq` | 35 | | | |
| _ | `(a.highSupportNoun:p\|dpTreeRelation) + p.form` | 116 | | | |
| _ | `(a.highSupportNoun:p\|dpTreeRelation) + p.spForm` | 118 | | | |
| _ | `(a.lowSupportNoun:p\|dpTreeRelation) + p.spPos` | 56 | | | |
| _ | `(a.lowSupportVerb:p\|dpTreeRelation) + a.form` | 105 | | | |
| _ | `(a.lowSupportVerb:p\|dpTreeRelation) + a.spForm` | 107 | | | |

Table 18: Feature templates of $S_{N+P}^l$ besides Tables 10 and 12

# References

Carreras, X., & Màrquez, L. (2005). Introduction to the CoNLL-2005 shared task: Semantic role labeling. In *Proceedings of the Ninth Conference on Computational Natural Language Learning*, pp. 152–164, Ann Arbor, Michigan.

Che, W., Li, Z., Hu, Y., Li, Y., Qin, B., Liu, T., & Li, S. (2008). A cascaded syntactic and semantic dependency parsing system. In *Proceedings of the Twelfth Conference on Computational Natural Language Learning*, pp. 238–242, Manchester.

Chen, S. F., & Rosenfeld, R. (1999). A Gaussian prior for smoothing maximum entropy models. Technical report CMU-CS-99-108, School of Computer Science, Carnegie Mellon University.

Chen, W., Kawahara, D., Uchimoto, K., Zhang, Y., & Isahara, H. (2008). Dependency parsing with short dependency relations in unlabeled data. In *Proceedings of the Third International Joint Conference on Natural Language Processing*, Vol. 1, pp. 88–94, Hyderabad, India.

Ciaramita, M., Attardi, G., Dell'Orletta, F., & Surdeanu, M. (2008). DeSRL: A linear-time semantic role labeling system. In *Proceedings of the Twelfth Conference on Computational Natural Language Learning*, pp. 258–262, Manchester.

Crammer, K., & Singer, Y. (2003). Ultraconservative online algorithms for multiclass problems. *The Journal of Machine Learning Research*, *3*(Jan), 951–991.

Dang, H. T., & Palmer, M. (2005). The role of semantic roles in disambiguating verb senses. In *Proceedings of the 43rd Annual Meeting of the Association for Computational Linguistics*, pp. 42–49, Ann Arbor, Michigan.

Ding, W., & Chang, B. (2008). Improving Chinese semantic role classification with hierarchical feature selection strategy. In *Proceedings of the 2008 Conference on Empirical Methods in Natural Language Processing*, pp. 324–333, Honolulu, Hawaii.

Gildea, D., & Jurafsky, D. (2002). Automatic labeling of semantic roles. *Computational Linguistics*, *28*(3), 245–288.

Hajič, J., Ciaramita, M., Johansson, R., Kawahara, D., Martí, M. A., Màrquez, L., Meyers, A., Nivre, J., Padó, S., Štěpánek, J., Straňák, P., Surdeanu, M., Xue, N., & Zhang, Y. (2009). The CoNLL-2009 shared task: Syntactic and semantic dependencies in multiple languages. In *Proceedings of the Thirteenth Conference on Computational Natural Language Learning: Shared Task*, pp. 1–18, Boulder, Colorado.

Henderson, J., Merlo, P., Musillo, G., & Titov, I. (2008). A latent variable model of synchronous parsing for syntactic and semantic dependencies. In *Proceedings of the Twelfth Conference on Computational Natural Language Learning*, pp. 178–182, Manchester.

Jiang, Z. P., & Ng, H. T. (2006). Semantic role labeling of NomBank: A maximum entropy approach. In *Proceedings of the 2006 Conference on Empirical Methods in Natural Language Processing*, pp. 138–145, Sydney.

Johansson, R., & Nugues, P. (2008). Dependency-based syntactic–semantic analysis with PropBank and NomBank. In *Proceedings of the Twelfth Conference on Computational Natural Language Learning*, pp. 183–187, Manchester.

Koo, T., Carreras, X., & Collins, M. (2008). Simple semi-supervised dependency parsing. In *Proceedings of the 46th Annual Meeting of the Association for Computational Linguistics: Human Language Technologies*, pp. 595–603, Columbus, Ohio.

Koomen, P., Punyakanok, V., Roth, D., & Yih, W.-T. (2005). Generalized inference with multiple semantic role labeling systems. In *Proceedings of the Ninth Conference on Computational Natural Language Learning*, pp. 181–184, Ann Arbor, Michigan.

Liu, C., & Ng, H. T. (2007). Learning predictive structures for semantic role labeling of NomBank. In *Proceedings of the 45th Annual Meeting of the Association of Computational Linguistics*, pp. 208–215, Prague.

Marcus, M. P., Santorini, B., & Marcinkiewicz, M. A. (1993). Building a large annotated corpus of English: The Penn Treebank. *Computational Linguistics, Special Issue on Using Large Corpora: II*, *19*(2), 313–330.

Màrquez, L., Surdeanu, M., Comas, P., & Turmo, J. (2005). A robust combination strategy for semantic role labeling. In *Proceedings of Human Language Technology Conference and Conference on Empirical Methods in Natural Language Processing*, pp. 644–651, Vancouver.

McDonald, R., & Pereira, F. (2006). Online learning of approximate dependency parsing algorithms. In *Proceedings of the Eleventh Conference of the European Chapter of the Association for Computational Linguistics*, pp. 81–88, Trento, Italy.

McDonald, R., Pereira, F., Ribarov, K., & Hajič, J. (2005). Non-projective dependency parsing using spanning tree algorithms. In *Proceedings of Human Language Technology Conference and Conference on Empirical Methods in Natural Language Processing*, pp. 523–530, Vancouver, British Columbia.

Meyers, A., Reeves, R., Macleod, C., Szekely, R., Zielinska, V., Young, B., & Grishman, R. (2004). The NomBank project: An interim report. In Meyers, A. (Ed.), *HLT-NAACL 2004 Workshop: Frontiers in Corpus Annotation*, pp. 24–31, Boston.

Meza-Ruiz, I., & Riedel, S. (2009). Jointly identifying predicates, arguments and senses using Markov logic. In *Proceedings of Human Language Technologies: The 2009 Annual Conference of the North American Chapter of the Association for Computational Linguistics*, pp. 155–163, Boulder, Colorado.

Nash, S. G., & Nocedal, J. (1991). A numerical study of the limited memory BFGS method and truncated-Newton method for large scale optimization. *SIAM Journal on Optimization*, *1*(2), 358–372.

Nivre, J., & McDonald, R. (2008). Integrating graph-based and transition-based dependency parsers. In *Proceedings of the 46th Annual Meeting of the Association for Computational Linguistics: Human Language Technologies*, pp. 950–958, Columbus, Ohio.

Nocedal, J. (1980). Updating quasi-Newton matrices with limited storage. *Mathematics of Computation*, *35*(151), 773–782.

Palmer, M., Gildea, D., & Kingsbury, P. (2005). The Proposition Bank: An annotated corpus of semantic roles. *Computational Linguistics*, *31*(1), 71–106.

Pradhan, S., Ward, W., Hacioglu, K., Martin, J., & Jurafsky, D. (2005). Semantic role labeling using different syntactic views. In *Proceedings of the 43rd Annual Meeting of the Association for Computational Linguistics*, pp. 581–588, Ann Arbor, Michigan.

Pradhan, S. S., Ward, W. H., Hacioglu, K., Martin, J. H., & Jurafsky, D. (2004). Shallow semantic parsing using support vector machines. In *Proceedings of the Human Language Technology Conference of the North American Chapter of the Association for Computational Linguistics*, pp. 233–240, Boston.

Punyakanok, V., Roth, D., Yih, W., & Zimak, D. (2004). Semantic role labeling via integer linear programming inference. In *Proceedings of the 20th International Conference on Computational Linguistics*, pp. 1346–1352, Geneva.

Pustejovsky, J., Meyers, A., Palmer, M., & Poesio, M. (2005). Merging PropBank, NomBank, TimeBank, Penn Discourse Treebank and coreference. In *Proceedings of the Workshop on Frontiers in Corpus Annotations II: Pie in the Sky*, pp. 5–12, Ann Arbor, Michigan.

Riedel, S. (2008). Improving the accuracy and efficiency of map inference for markov logic. In *Proceedings of the Twenty-Fourth Conference Annual Conference on Uncertainty in Artificial Intelligence*, pp. 468–475, Corvallis, Oregon.

Roth, D., & Yih, W. (2004). A linear programming formulation for global inference in natural language tasks. In *Proceedings of the Eighth Conference on Computational Natural Language Learning*, pp. 1–8, Boston.

Surdeanu, M., Johansson, R., Meyers, A., Màrquez, L., & Nivre, J. (2008). The CoNLL 2008 shared task on joint parsing of syntactic and semantic dependencies. In *Proceedings of the Twelfth Conference on Computational Natural Language Learning*, pp. 159–177, Manchester.

Surdeanu, M., Marquez, L., Carreras, X., & Comas, P. R. (2007). Combination strategies for semantic role labeling. *Journal of Artificial Intelligence Research*, *29*, 105–151.

Titov, I., Henderson, J., Merlo, P., & Musillo, G. (2009). Online graph planarisation for synchronous parsing of semantic and syntactic dependencies. In *Proceedings of the 21st International Jont Conference on Artifical Intelligence*, pp. 1562–1567, Pasadena, California.

Toutanova, K., Haghighi, A., & Manning, C. D. (2005). Joint learning improves semantic role labeling. In *Proceedings of the 43rd Annual Meeting on Association for Computational Linguistics*, pp. 589–596, Ann Arbor, Michigan.

Xue, N. (2006). Semantic role labeling of nominalized predicates in Chinese. In *Proceedings of Human Language Technology Conference of the North American Chapter of the Association of Computational Linguistics, Main Conference*, pp. 431–438, New York.

Xue, N., & Palmer, M. (2004). Calibrating features for semantic role labeling. In *Proceedings of the 2004 Conference on Empirical Methods in Natural Language Processing*, pp. 88–94, Barcelona.

Zhao, H., Chen, W., Kazama, J., Uchimoto, K., & Torisawa, K. (2009). Multilingual dependency learning: Exploiting rich features for tagging syntactic and semantic dependencies. In *Proceedings of the Thirteenth Conference on Computational Natural Language Learning: Shared Task*, pp. 61–66, Boulder, Colorado.

Zhao, H., Chen, W., Kit, C., & Zhou, G. (2009). Multilingual dependency learning: A huge feature engineering method to semantic dependency parsing. In *Proceedings of the Thirteenth Conference on Computational Natural Language Learning: Shared Task*, pp. 55–60, Boulder, Colorado.

Zhao, H., & Kit, C. (2008). Parsing syntactic and semantic dependencies with two single-stage maximum entropy models. In *Proceedings of the Twelfth Conference on Computational Natural Language Learning*, pp. 203–207, Manchester.